\documentclass[a4paper,conference]{IEEEtran}
\usepackage{ifpdf}
\usepackage{cite}
\usepackage{amsmath,bm}
\usepackage{algorithmic}
\usepackage{array}
\usepackage{url}
\usepackage{amssymb}
\usepackage{booktabs,color,threeparttable}
\usepackage{tabularx}
\usepackage{multirow}
\usepackage{tabulary}
\ifCLASSOPTIONcompsoc
 \usepackage[caption=false,font=normalsize,labelfont=sf,textfont=sf]{subfig}
\else
 \usepackage[caption=false,font=footnotesize]{subfig}
\fi
\ifCLASSINFOpdf
   \usepackage[pdftex]{graphicx}
   \graphicspath{{./figures/}}
   \DeclareGraphicsExtensions{.pdf,.jpeg,.png}
\else
   \usepackage[dvips]{graphicx}
   \graphicspath{{./eps/}}
   \DeclareGraphicsExtensions{.eps}
\fi

\begin{document}
%
\title{Feature Transformation for Cross-domain Few-shot Remote Sensing Scene Classification}



%
\author{\IEEEauthorblockN{Qiaoling Chen\IEEEauthorrefmark{1}, Zhihao Chen\IEEEauthorrefmark{1} and Wei Luo\IEEEauthorrefmark{1}\IEEEauthorrefmark{2}}
\IEEEauthorblockA{\IEEEauthorrefmark{1}College of Mathematics and Informatics\\
South China Agricultural University,
Guangzhou, GD, China 510000\\ Email:
chenql\_716@163.com, 202025710106@stu.scau.edu.cn
}
\IEEEauthorblockA{\IEEEauthorrefmark{2}Pazhou Lab, Guangzhou, China 510330,.
\\
Email: cswluo@gamil.com}
}

\maketitle

\begin{abstract}
Effectively classifying remote sensing scenes is still a challenge due to the increasing spatial resolution of remote imaging and large variances between remote sensing images. Existing research has greatly improved the performance of remote sensing scene classification (RSSC). However, these methods are not applicable to cross-domain few-shot problems where target domain is with very limited training samples available and has a different data distribution from source domain. To improve the model's applicability, we propose the feature-wise transformation module (FTM) in this paper. FTM transfers the feature distribution learned on source domain to that of target domain by a very simple affine operation with negligible additional parameters. Moreover, FTM can be effectively learned on target domain in the case of few training data available and is agnostic to specific network structures. Experiments on RSSC and land-cover mapping tasks verified its capability to handle cross-domain few-shot problems. By comparison with directly finetuning, FTM achieves better performance and possesses better transferability and fine-grained discriminability. \textit{Code will be publicly available.} 
\end{abstract}


\IEEEpeerreviewmaketitle

\section{Introduction}
\label{secI}
Remote sensing scene classification (RSSC) has attracted much attention in the field of optical remote sensing image processing and analysis in recent years, both due to the availability of high spatial-resolution images and the key role in wide applications, e.g., disaster detection~\cite{Huang2019MonitoringES}, environmental monitoring~\cite{Alcntara2012MappingAA}, urban planning~\cite{gid}. However, effectively classifying scenes from a newly obtained remote sensing image (RSI) is still nontrivial owing to the rich content brought by high-resolution, imaging conditions, seasonal changes and so on. Together with the difficulty of collecting sufficient labeled training samples, these factors make the robust-performance of RSSC a very challenging task. 

To improve the performance of RSSC, deep learning methods~\cite{alexnet12hiton,googlenet15Szegedy,resnet16kaiming} have been widely employed in RSSC. The deep learning based RSSC methods made use of the hierarchical network structure and feature abstraction ability of deep models to extract robust features for classification~\cite{Li2017IntegratingMF,Cao2021SelfAttentionBasedDF,Nogueira2017TowardsBE} and achieved a great success, although they usually set aside the distribution differences between the training and testing data. While in a more realistic setting, the distribution difference was explicitly taken into consideration (under the framework of domain adaption) to build more applicable RSSC models like~\cite{Zhu2022AttentionBasedMR,Othman2017DomainAN,Lu2020MultisourceCN}. These methods usually require the same class distribution in the source and target domains. In addition, existing methods are almost all built on the prerequisite that sufficient training samples are available on target domain. This is, however, a very strict constraint on many real RSSC applications, especially in those target samples from a different distribution.

\begin{figure}[t]
    \centering
    \includegraphics[width=\linewidth]{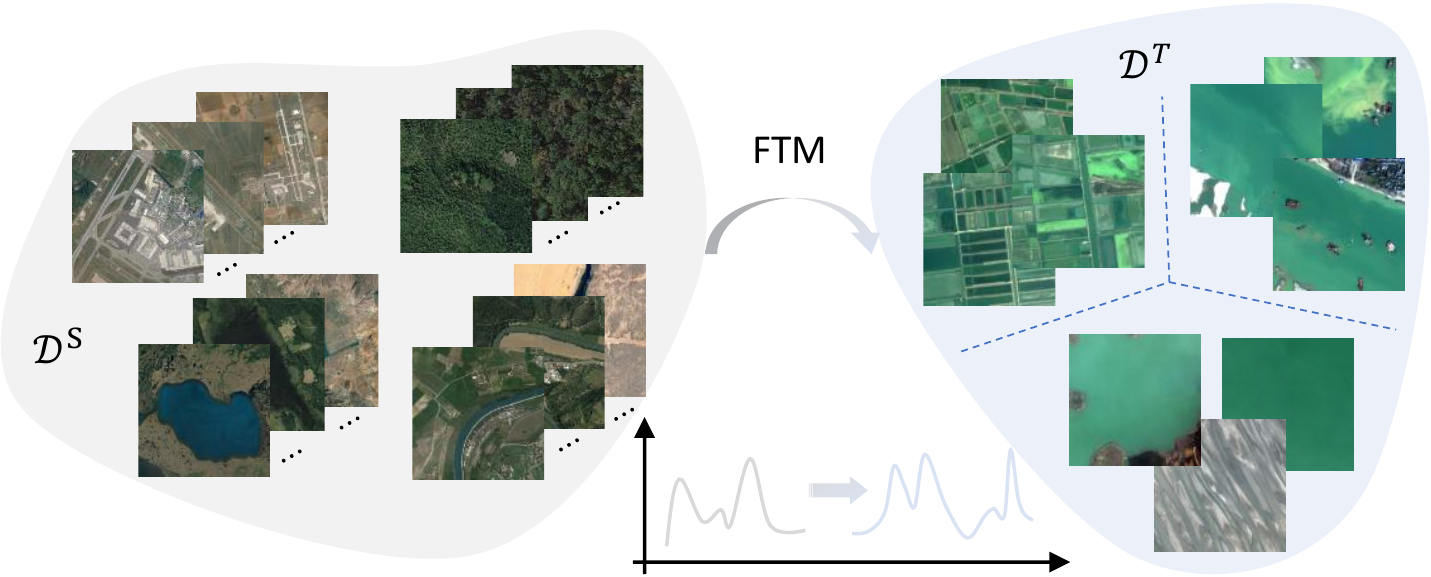}
    \caption{Illustration of the motivation for the proposed method. Source domain $\mathcal{D}^S$ has sufficient training samples for each class as shown here airplane, forest, lake, and river. Target domain $\mathcal{D}^T$ may have different classes from $\mathcal{D}^S$ and provides only few training samples for each class (here 3 samples for paddy field, river, and lake, respectively). As shown here, $\mathcal{D}^S$ and $\mathcal{D}^T$ have a significant domain gap. The proposed FTM tries to transfer the feature distribution learned on $\mathcal{D}^S$ to matching that of $\mathcal{D}^T$ by an affine transformation with a negligible number of additional parameters, thus improving the applicability of models learned on $\mathcal{D}^S$ to cross-domain few-shot tasks.}
    \label{fig:sketch}
\end{figure}

To address the difficulty of cross-domain RSSC tasks with few training samples, we propose a feature-wise transformation module (FTM) in deep CNNs with a two-stage training strategy. FTM borrows the idea from feature-wise linear modulation (FiLM)~\cite{film@aaai} but works in the unconditional setting and can be inserted in every convolutional layer. It attacks the cross-domain problem by transforming the distribution of features learned on source domain into matching that of target domain (see Fig~\ref{fig:sketch}). To achieve this, a pair of scale and shift vectors is applied to convolutional layers element-wisely. This pair of vectors, however, is not learned on source domain with the backbone network parameters, but instead trained on target domain without touching those already learned backbone parameters on source domain, which is different from~\cite{film@aaai,Tseng2020CrossDomainFC,tadam@nips} where the FiLM parameters are learned with the backbone network in an end-to-end manner. This two-stage training strategy can also alleviate the phenomenon of overfitting on target tasks with few labeled training samples due to the parsimonious parameters involved in the second training stage. Generally, the separated training strategy and the parsimonious usage of parameters in FTM make it well adapted to scenarios with limited labeled training samples and class distribution mismatching between domains. We compare FTM with directly finetuning in this study and show its better prediction performance, transferability, and fine-grained discriminability. We notice that there is no existing work to deal with this problem in RSSC and we approach this problem in this study with the following contributions:
\begin{itemize}
\item We propose FTM for cross-domain few-shot RSSC. FTM transforms the feature distribution of source data into those matching the target data via an affine transformation.
\item We propose a two-stage training strategy in which only FTM parameters are involved in the second training stage on target task, thus alleviating the overfitting problem.
\item We validate the effectiveness of FTM on a constructed cross-domain few-shot dataset in RSSC and demonstrate its applicability to land cover mapping tasks.
\end{itemize}

\section{Related Work}
\label{secII}
\textbf{Remote sensing scene classification} (RSSC) has gained great progress in recent years since the publication of several benchmark datasets such as AID~\cite{aid} and NWPU~\cite{nwpu}, which promotes the application of deep models in RSSC. In the early studies, researches focus on directly transferring deep features~\cite{Nogueira2017TowardsBE} or exploring deep network structures to utilize multi-layer~\cite{He2018RemoteSS,Lu2019AFA,Li2017IntegratingMF} or multi-scale features~\cite{Liu2018LearningMD,Liu2018SceneCB,Wang2021EnhancedFP}, thus fully exploiting granularity information in RSIs~\cite{Wang2020MultiGranularityCA}. 
Another line of research highlights the importance of local structures and geometries and proposes to combine them with global features for more discriminative representation~\cite{Yuan2019RemoteSI,Li2020DeepMI,FengpengLi2020HighResolutionRS}. Recently, the attention mechanism is further incorporated to selectively attend informative areas~\cite{optimal} or assign objects with different weights for feature fusion~\cite{Cao2021SelfAttentionBasedDF}. In addition, nonlocal attentions are also studied to integrate long-range spatial relationship for RSSC~\cite{Thrifty}. Although the mainstream deep learning methods are 
absorbed quickly by the RSSC field and much progress has been achieved, these methods, however, are not applicable to our setting in this paper where the training and testing data has different distributions.

\textbf{Few-shot learning} (FSL) has attracted much attention in recent years where the target tasks have very few training samples available. To tackle this problem, three kinds of methodologies are usually employed. The metric-learning based methods~\cite{matchingnet,protonet,relationnet} target at learning an embedding space where an off-the-shelf or learned metric can perform well. In contrast, the meta-learning based methods~\cite{maml,optnet,metaoptnet} aim to make the learned model can fast adapt to unseen novel tasks at the test stage. Recently, the finetuning based methods~\cite{closelook} report exciting results by exploiting multiple subspaces~\cite{tafssl} or assembling multiple CNN features~\cite{Chowdhury2021FewshotIC}. Meanwhile, FSL is also developed in settings like incremental learning~\cite{topic,aan}, cross-domain~\cite{Tseng2020CrossDomainFC,Phoo2021SelftrainingFF}, etc. However, very few works investigate FSL in RSSC while it is the core problem in this study.

\textbf{Domain adaption} (DA) has gone through thorough studies and been introduced into RSSC for a long time. The research of DA in RSSC mainly borrows ideas of existing DA approaches such as by finetuning models on target domain~\cite{gid}, by minimizing the maximum mean discrepancy between the source and target data distributions~\cite{Othman2017DomainAN}. ~\cite{Zhu2022AttentionBasedMR} proposes combining the marginal and conditional distributions for more comprehensive alignment, AFGAN~\cite{advfg} captures structures behind data and local information for fine-grained alignment. In addition, the class distribution misaligned problem is investigated in~\cite{Lu2020MultisourceCN} by multisource compensation learning. Nevertheless, these methods assume sufficient training samples available on target domain.~\cite{Yan2019CrossDomainDM} studies the cross-domain task with limited target samples in RSSC, their training samples on the target domain is, however, orders of magnitude larger than our's.  

\section{Approaches}
\label{secIII}
In this section, we propose FTM in deep CNNs that adapts the feature distribution learned on source domain to that of target domain. Assuming a well-labeled large-scale dataset and a newly acquired RS image with a small number of labeled samples annotated from it, we define two domains, the source domain $\mathcal{D}^S$ and the target domain $\mathcal{D}^T$, respectively. The data of the two domains can from different classes, $\mathcal{C}^S \neq \mathcal{C}^T$ and $\mathcal{C}^S\cap \mathcal{C}^T \neq \O$. Our approach first learns a backbone network on $\mathcal{D}^S$, and then adapts the backbone feature maps by FTM on $\mathcal{D}^T$ without touching the backbone network parameters. In the following, we start by introducing FTM, followed by describing its training strategy and then present the FTM network.



\begin{figure*}
    \centering
    \includegraphics[width=0.8\linewidth]{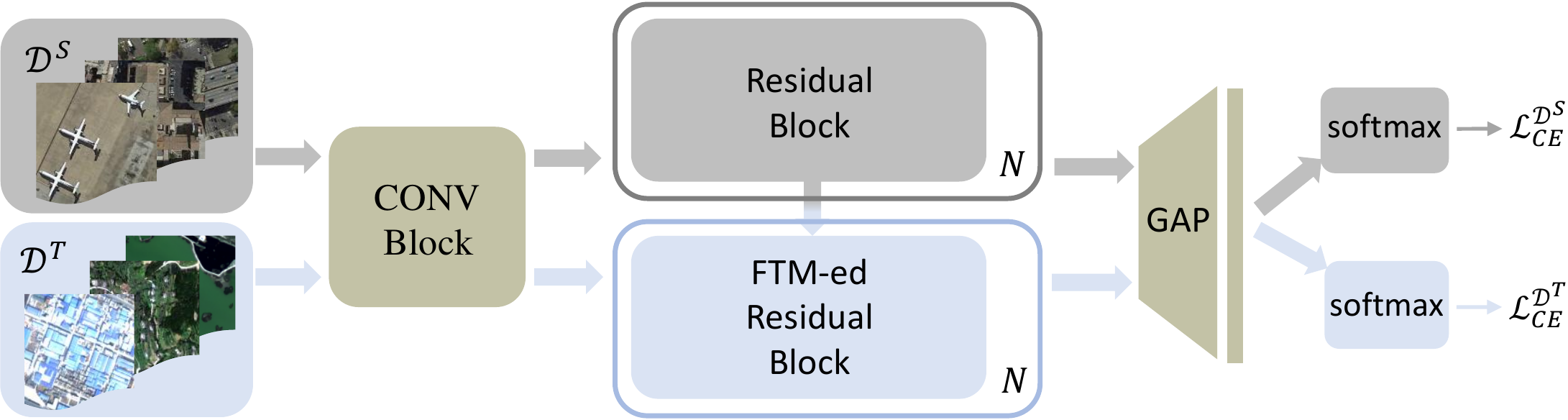}
    \caption{Overview of the proposed FTM network. The detail of the FTM-ed residual block is depicted in Fig.~\ref{fig:ftmblock}(b). Our approach first trains a backbone network (shaded by gray blocks) on the source domain $\mathcal{D}^S$ and then uses it to initialize a corresponding FTM network (shaded by blue blocks). The aligned parts between the two networks are then fixed and only the remained parts of the FTM network are learned on the target domain $\mathcal{D}^T$ by $\mathcal{L}^{\mathcal{D}^T}_{CE}$. The light green blocks are shared. Best viewed in color.}
    \label{fig:overview}
\end{figure*}
\subsection{Feature-wise Transformation Module}
\label{sec:ftm}
Modern deep CNNs usually include BN~\cite{bn@icml} layers that reduce internal covariate shift and preserve feature distributions via a learned affine transformation for training efficiency. This operation inspires us to model different feature distributions by adjusting the feature maps activations of a learned CNN, expecting it can perform well on a different domain with few training examples.

Supposing a backbone network has been trained on $\mathcal{D}^S$. Feature-wise transformation module (FTM) transforms the feature maps by a pair of scale and shift vectors $(\bm{\gamma},\bm{\beta})$. Concretely, assuming the feature maps of an input $\bm X\in\mathbb{R}^{3\times H\times W}$from the $l$-th layer is $\bm f^l\in \mathbb{R}^{C\times H'\times W'}$, FTM transforms the distribution of $\bm f^l$ by modulating its activations:
\begin{equation}
    \tilde{\bm f}^l_c = \bm\gamma_c^l \odot \bm f^l_c + \bm\beta^l_c,
\end{equation}
where the subscript $c$ represents feature channel indices and $\odot$ means element-wise multiplication, $\bm \gamma^l,\bm\beta^l\in\mathbb{R}^C$ are learnable parameters. FTM approaches the distribution change of $\bm f^l$ by independently changing each feature map's activations. Compared to FiLM~\cite{film@aaai}, where $(\bm{\gamma},\bm{\beta})$ are generated by a conditioning network, FTM works in a unconditional setting and simply initializes $\bm\gamma$ and $\bm\beta$ to $\bm 1$ and $\bm 0$, respectively, and let it adapt with the learning on target domain. By noting that the BN transform recovers feature activations through an affine operation, FTM further adapts it to a large range and recovers the BN transform at $\bm\gamma=\bm 1$ and $\bm\beta=\bm 0$. This simplification not only benefits the optimization of FTM on few-shot tasks but also preserves the properties of FiLM. 

\subsection{Optimization}
\label{sec:optimization}
To alleviate the overfitting phenomenon of deep CNNs with FTM on target domain with few labeled training samples, we study a two-stage learning strategy for optimization. 
Recalling that our target is transforming the feature distribution learned on source domain into that of target domain, we prefer to keep the backbone parameters unchanged and only train FTM on target data. To this end, we first optimize the backbone network by regular training on $\mathcal{D}^S$, then we fix the backbone network parameters and optimize FTM parameters $\{\bm\gamma,\bm\beta\}$ on $\mathcal{D}^T$ through SGD. 

Intuitively, we put FTM between the BN layers and nonlinear activations. This operation, however, will cause the shift of mid-level feature activations if we keep the backbone network parameters untouched, thus complicating optimization. To this end, we free the statistics of BN layers by making them adapt to input changes, and leave the shift in activations to be compensated by $\{\bm\gamma,\bm\beta\}$.  

\subsection{The FTM Network}
\label{sec:ftmnetwork}
\begin{figure}
    \centering
    \begin{minipage}{0.4\linewidth}
    \includegraphics[width=0.75\linewidth]{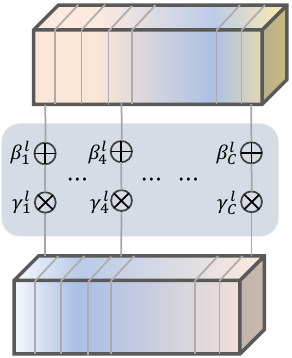}\\
    \centering (a)
    \end{minipage}
    \begin{minipage}{0.4\linewidth}
    \includegraphics[width=0.6\linewidth]{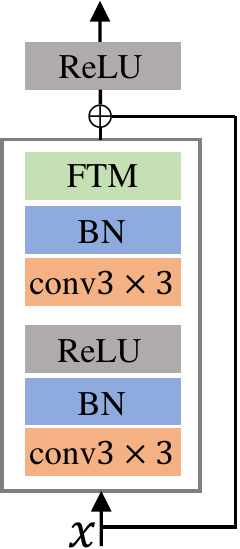}\\
    \centering (b)
    \end{minipage}
    \caption{(a) The shaded part is FTM, which operates on feature maps channel-wise. $\otimes$ and $\oplus$ represent element-wise multiplication and addition. (b) A FTM-ed residual block.}
    \label{fig:ftmblock}
\end{figure}

We instantiate our FTM network on the backbone of ResNet-34~\cite{resnet16kaiming}. It is worth noting that FTM is agnostic to specific CNN structures and we choose ResNet-34 just for simplicity. ResNet-34 includes one convolutional stem and 4 stages each with several residual blocks. Each residual block has two convolutional layers to form a shortcut connection. We construct the corresponding FTM network by inserting FTM after the BN layer of the second convolutional layer of the last residual block in one or several stages. For simplicity, we insert FTM after the BN layer of conv5\_3 in ResNet-34 to illustrate its strength in this work. The transformed feature maps are then rectified by ReLU~\cite{hinton10rectified_active_function} and globally averaged pooled to be fed into a softmax function for classification. Fig.~\ref{fig:ftmblock} shows the FTM-ed residual bock in conv5. 

\section{Experiments}
\label{secIV}
In this section, we evaluate the transferability of the FTM network on two cross-domain few-shot applications: a RSSC task and a land-cover mapping task.

\subsection{Datasets}
Two datasets from different imaging conditions are collected as the source and target domains, respectively. The source domain data are from NWPU-RESISC45~\cite{nwpu}, which has $700$ of size $256\times 256$ Google Earth RGB images for each class with a total of 45 scene classes such as residential areas, basketball courts, and commercial areas.
The target domain data are from the R, G, and B channels of GID~\cite{gid} multispectral images, which are collected from Gaofen-2 satellite with a spatial resolution of 4m. GID provides two subsets -- a large-scale classification set (Set-C) and a fine land-cover classification set (Set-F). Set-C includes $150$ and $30$ training and validation images of size $6800\times 7200$ with each pixel annotated into 5 coarse categories. Set-F has a subset of $6,000$ image patches with train/val/testing $1500/3750/750$ respectively. The image patches are of size $224\times 224$ and belong to 15 fine categories, which are subcategories of the 5 coarse categories. Set-F is used as $\mathcal{D}^T$ and images from Set-C are only used for land-cover mapping evaluation. We report the average performance over $5$ trials on the RSSC task. 

\subsection{Implementation}
We experiment with a FTM network based on the ResNet-34 backbone. The ResNet-34 pretrained on ImageNet~\cite{imagenet@feifei} is first employed to learn on $\mathcal{D}^S$, where random crops of size $224\times 224$ are used for training and $100$ images from each class are kept for validation. We train ResNet-34 by Adam~\cite{Kingma2015AdamAM} on $\mathcal{D}^S$ for $30$ epochs with batch size $128$, \textit{lr} $10^{-4}$, and decay \textit{lr} by 0.1 every 10 epochs. After this stage, we select the best-performed one to initialize the FTM network, keep the aligned parameters fixed, and learn the remained parameters on $\mathcal{D}^T$ for the RSSC task. The learning hyper-parameters are presented in Table~\ref{tab:params}. For the land-cover mapping task, we classify every pixel into one of the 5 coarse classes by combining the output probabilities of subcategories that belong to the same coarse category. 

\textit{Baseline}: we compare FTM network with the finetuning (FT) method, which directly finetunes the best-performed ResNet-34 trained on $\mathcal{D}^S$ on $\mathcal{D}^T$. The finetuning hyper-parameters are in Table~\ref{tab:params}

\begin{table}
\begin{center}
\begin{threeparttable}
\caption{Learning hyper-parameters of FTM and FT on $\mathcal{D}^T$.}
\label{tab:params}
\begin{tabular*}{\linewidth}{@{}@{\extracolsep{\fill}}cccccccc@{}}
\toprule
    & batch & epochs & lr & step & decay &opt  \\
\midrule
FT & 64 &50 &0.001  &15  &0.1  &Adam \\
\midrule
FTM  & 64 &50 &0.003  &15  &0.1  &Adam\\
\bottomrule
\end{tabular*}
\end{threeparttable}
\end{center}
\end{table}


\subsection{Experimental Results}
\begin{figure}
    \centering
    \includegraphics[width=\linewidth]{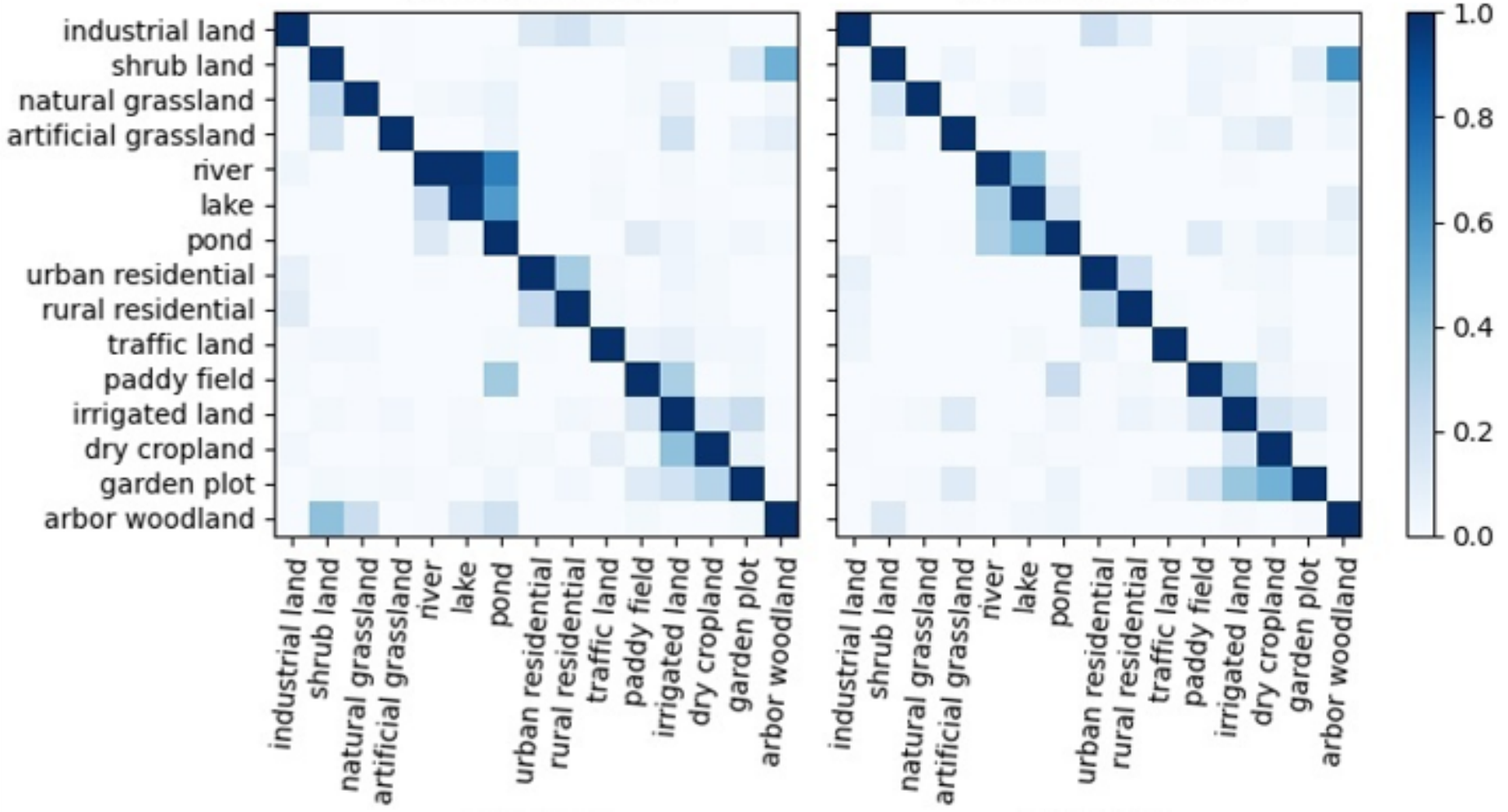}
    \caption{The confusion matrices of finetuning (left) and FTM (right) networks on the testing set of Set-F. Both networks are trained with 10 shots on Set-F.}
    \label{fig:confusion}
\end{figure}


\textbf{RSSC Results}. 
Table~\ref{tab:different-shot} compares the performance of finetuning and FTM under various cross-domain few-shot settings. The results are obtained from the Set-F testing set with models trained on different shots ranging from 3 to 50, and show that FTM improves the performance over finetuning by $3.1\%$ on average, demonstrating the clear advantages of FTM. In addition, Table~\ref{tab:different-shot} illustrates that the performance of both FTM and finetuning can be steadily improved with more training shots and the improvement of FTM over finetuning is relatively stable independent of the number of available training shots. These observations validate that FTM possesses the ability to transform the feature distribution learned on $\mathcal{D}^S$ into that of target domain even with very limited training shots available on the target domain, thus alleviating the tendency to overfitting on the target domain.

To better understand which aspects of advantages brought by FTM, we make an analysis of the confusion matrices of FTM and finetuning networks trained on $10$ shots in Fig.~\ref{fig:confusion}. It can be seen that FTM has a more concentrated diagonal distribution than finetuning, indicating its better classification performance, especially in those subcategories belonging to the same coarse category. Specifically, we find that FTM can well separate arbor woodland from shrub land and distinguish river, lake, and pond effectively, which are respectively from the same coarse categories -- forest and water, and confused by the finetuning method. This signifies that FTM has the ability to transform the original feature space into a more delicate and discriminative space where the subtle differences between fine-grained categories can be better ascertained, even in the case of very limited training shots available.


\begin{table}
\begin{center}
\begin{threeparttable}
\caption{Accuracy of fine-tuning (FT) and FTM method on different shots. The standard deviations are within $0.03$.}
\label{tab:different-shot}
\begin{tabular*}{\linewidth}{@{}@{\extracolsep{\fill}}cccccccc@{}}
\toprule
    & 3 & 5 & 10 & 15 & 20 & 30 & 50  \\
\midrule
FT & 0.50 & 0.57 & 0.65 & 0.71 & 0.73 & 0.73 & 0.81 \\
\midrule
FTM  & 0.53 & 0.59 & 0.69 & 0.73 & 0.77 & 0.77 & 0.84 \\
\bottomrule
\end{tabular*}
\end{threeparttable}
\end{center}
\end{table}


\begin{figure*}[t]
    \centering
    \includegraphics[width=0.8\linewidth]{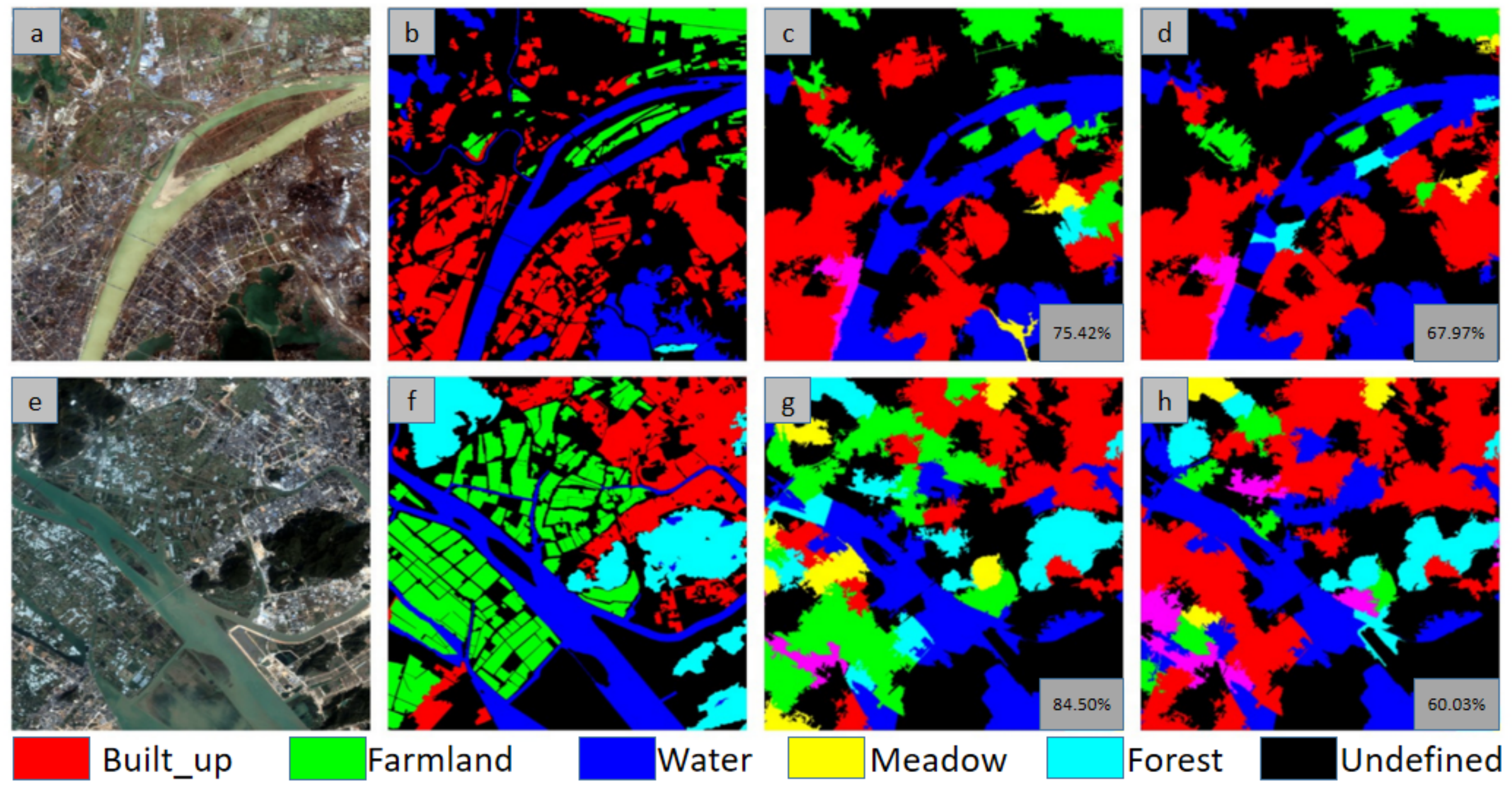}
    \caption{The land-cover mapping results. (a) and (e) are RGB images from GID validation set. (b) and (f) are ground-truth annotations. (c, g) and (d, h) are mapping results from FTM and finetuning, respectively. The numbers at the bottom of (c, d, g, h) are class average F1 scores evaluated on $224\times 224 $ image patches.}
    \label{fig:mapping}
\end{figure*}

\textbf{Land Cover Mapping Results}. 
To verify that FTM can improve models applicability to across-domain tasks, we perform the land-cover mapping task on two randomly selected GID images from the Set-C validation set. The two GID images are taken from different locations and seasons shown a big domain gap to the images in $\mathcal{D}^S$. For simplicity, we do not annotate additional training samples from the two GID images as the target domain data but directly use the Set-F training samples as target domain data since they are obtained from the same satellite. To achieve pixel-level mapping, we on the one hand segment the full GID image into $224\times 224$ patches and classify them by using the FTM (or finetuning) networks, on the other hand, we segment it into $100$ superpixels by using SLIC~\cite{Achanta2012SLICSC} and align them with the $224\times 224$ patches. Finally, we assign labels to superpixels by assembling the labels of $224\times 224$ patches within the corresponding superpixels and labeling it by winner-take-all.

\begin{table}
\caption{F1 scores ($\%$) of finetuning (FT) and FTM networks on land-cover mapping tasks with 3 training samples each class.}
\label{tab:land-cover-3}
\centering
\begin{threeparttable}
\begin{tabular*}{0.6\columnwidth}{@{}@{\extracolsep{\fill}}lcc@{}}
\toprule
					&FT3 	& FTM3 	\\
\midrule
Farmland &55.3 & 86.3\\
Built-up &80.6 & 90.0\\
Forest &35.2 & 53.0\\
Water &84.8 & 90.5\\
\midrule
Average &64.0 & 80.0\\
\bottomrule
\end{tabular*}
\end{threeparttable}
\end{table}

Table~\ref{tab:land-cover-3} shows the average F1 scores of finetuning and FTM networks evaluated on the $224\times 224$ patches of the two GID images. By comparison, FTM shows clear advantage over finetuning, achieving higher performance on all categories. Noting that there is no meadow class because the image has no pixels belonging to it. Further, it is worth special attention that the improvement on farmland is very significant raising from $55.3\%$ to $\mathbf{86.3\%}$. These improvements further validate the wide applicability of FTM to cross-domain few-shot tasks considering that we even do not annotate training samples from the target image.

We further visualize the mapping results in Fig.~\ref{fig:mapping}. From it we can find that GID images have large variances between them. This poses great obstacles to model's applicability where large number of annotated training samples are usually needed to retrain the model. However, FTM can alleviate the annotation requirements. The third and fourth columns of Fig.~\ref{fig:mapping} show prediction results. By comparison, we conclude that FTM can effectively predict main areas in the image and keep the smoothness between neighboring superpixels. In contrast, finetuning fails to achieve these effects and results in fragmented superpixels. 
For example, large areas of farmland are mismapped into built-up by finetuning while correctly mapped by FTM. 
This is because seasonal changes cause large differences between the source and target domains in the farmland class, 
thus when the labeling information of the target data is limited, it is incapable of the finetuning method to effectively represent contextual properties of this scene class.
Although the visualization effects are far behind satisfaction, we, however, should note that our purpose is to validate the transferability of FTM across domains while not the mapping accuracy, which can be achieved via much smaller image patches and more superpixels.




\section{Conclusion}
\label{secV}
In this paper, we studied a feature-wise transformation module (FTM) that transforms feature distributions learned on the source domain into that of target domain. FTM can quickly adapt to target domain with very limited training data and effectively alleviate overfitting. Experiments on RSSC and land-cover mapping tasks verified its transferability, fine-grained discriminability, and illustrate its advantages over the finetuning method, especially in those cases with very limited training shots available. Although FTM is simple, it shows great applicability to the RS field where large domain gaps exist and available training samples are extremely limited.



\section*{Acknowledgment}
This work was supported in part by NSFGD (No.2020A1515010813), STPGZ (No.202102020673), Young Scholar Project of Pazhou Lab (No.PZL2021KF0021), and NSFC (No.61702197),


%

\bibliographystyle{IEEEtran}
\bibliography{mybib}

\end{document}